\documentclass[journal,twoside,web]{ieeecolor}

\usepackage{generic}
\usepackage{color,array}
\usepackage{graphicx}
\usepackage{amsmath}
\usepackage{amssymb}
\usepackage{verbatim}
\usepackage{epstopdf}
\usepackage{tabularx} 
\usepackage{booktabs}   
\usepackage{mathrsfs} 
\usepackage{multirow} 
\usepackage{amsfonts} 
\usepackage{wrapfig}  
\usepackage{cite}

\def\0{{\bf 0}}
\def\1{{\bf 1}}

\def\etal{{\em et al.}}
\def\eg{{\em e.g.}}
\def\ie{{\em i.e.}}

\def\etal{{\em et al.\/}\,}

\usepackage[colorlinks=true,
            linkcolor=blue,
            citecolor=blue,
            urlcolor=blue]{hyperref} 

\begin{document}

\title{Detecting Performance Degradation under Data Shift in Pathology Vision-Language Model}

\author{Hao~Guan,  
\and
    Li~Zhou
\thanks{H.~Guan and L.~Zhou are with the Division of General Internal Medicine and Primary Care, Brigham and Women’s Hospital, and Department of Medicine, Harvard Medical School, Boston, Massachusetts 02115, USA.  
Corresponding Author: H.~Guan (hguan6@bwh.harvard.edu).} 
}

\markboth{IEEE XXX, Vol. XX, No. XX, JANUARY 2026}
{Hao Guan \MakeLowercase{\textit{et al.}}: Detecting Performance Degradation under Data Shift in Pathology Vision-Language Model}

\maketitle
\begin{abstract}
Vision-Language Models have demonstrated strong potential in medical image analysis and disease diagnosis. However, after deployment, their performance may deteriorate when the input data distribution shifts from that observed during development. Detecting such performance degradation is essential for clinical reliability, yet remains challenging for large pre-trained VLMs operating without labeled data.
In this study, we investigate performance degradation detection under data shift in a state-of-the-art pathology VLM. We examine both input-level data shift and output-level prediction behavior to understand their respective roles in monitoring model reliability. To facilitate systematic analysis of input data shift, we develop \emph{DomainSAT}, a lightweight toolbox with a graphical interface that integrates representative shift detection algorithms and enables intuitive exploration of data shift. Our analysis shows that while input data shift detection is effective at identifying distributional changes and providing early diagnostic signals, it does not always correspond to actual performance degradation.
Motivated by this observation, we further study output-based monitoring and introduce a label-free, confidence-based degradation indicator that directly captures changes in model prediction confidence. We find that this indicator exhibits a close relationship with performance degradation and serves as an effective complement to input shift detection.
Experiments on a large-scale pathology dataset for tumor classification demonstrate that combining input data shift detection and output confidence-based indicators enables more reliable detection and interpretation of performance degradation in VLMs under data shift. These findings provide a practical and complementary framework for monitoring the reliability of foundation models in digital pathology.
\end{abstract}



\section{Introduction}
\IEEEPARstart{R}{ecent} advances in Vision-Language Models (VLMs) have greatly expanded the capabilities of artificial intelligence in medicine~\cite{radford2021learning,
wang2022medclip,li2023llava,Li2025Exploring}. By jointly encoding visual and textual information, VLMs (\eg, CLIP~\cite{radford2021learning}) and their medical adaptations have enabled flexible zero-shot classification, image-report retrieval, and caption generation across clinical imaging modalities. In digital pathology, VLMs offer a promising foundation for computational diagnosis in pathology.

\begin{figure}[t] 
\setlength{\belowcaptionskip}{-2pt}
\setlength{\abovecaptionskip}{-2pt}
\setlength{\abovedisplayskip}{-2pt}
\setlength{\belowdisplayskip}{-2pt}
\center
 \includegraphics[width= 1.0\linewidth]{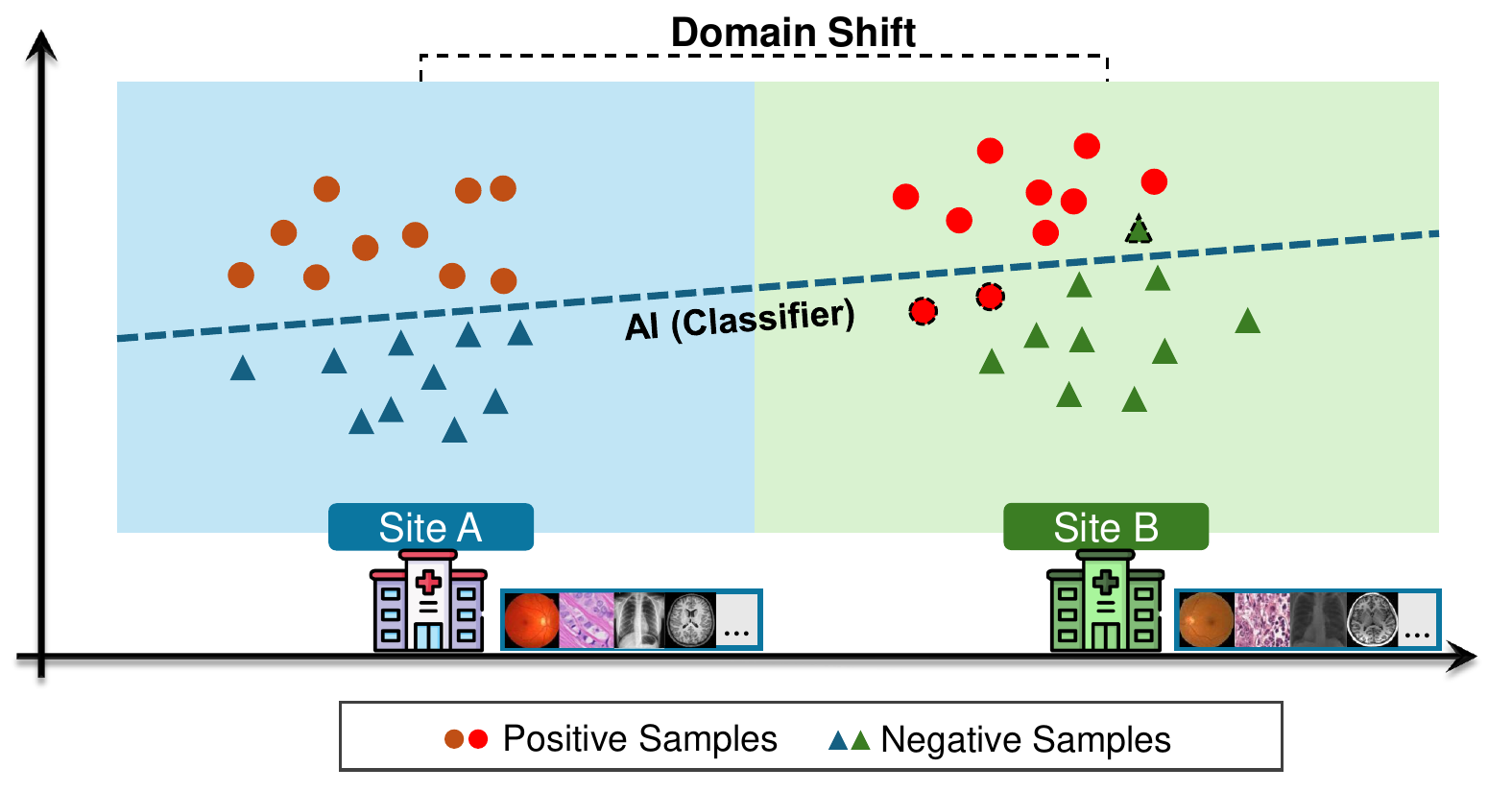}
 \caption{Overview of domain (data) shift across different medical sites and its impact on performance of medical AI models. An AI model deployed in Site A operates well, but when applied to Site B, the model suffers degradation under data shift.} 
 \label{fig_shift}
\end{figure}

Despite these advantages, the long-term reliability of VLMs in clinical deployment remains uncertain. Once deployed, these models may encounter input data that significantly differ from the observed distribution, arising from variations in scanners, staining procedures, or acquisition sites. This is the well-known domain (data) shift problem~\cite{finlayson2021clinician,stacke2020measuring,guan2021domain,guan2023domainatm}. Such data shifts, as shown in Fig.~\ref{fig_shift}, can cause performance degradation, compromising the safety and trustworthiness of AI-assisted diagnostic systems~\cite{guan2025keeping}. Detecting degradation is therefore critical to prevent clinical risks and guide model maintenance.

However, detecting performance degradation in large pre-trained pathology VLMs presents several fundamental challenges. First, despite the rapid adoption of VLMs in medical imaging, their post-deployment reliability, particularly performance degradation under real-world data shift, has not been systematically studied. 
Second, in practical medical settings, ground-truth labels are often unavailable after deployment, making direct detection of performance degradation using standard metrics such as accuracy or AUC infeasible.

To address this challenge, we investigate performance degradation under data shift in a state-of-the-art pathology VLM by jointly analyzing input data shift and output-level prediction. To support systematic exploration of input data shift, we develop \textbf{DomainSAT}, a lightweight GUI-based toolbox that integrates representative shift-detection algorithms, enabling intuitive inspection of data shifts across datasets. Our analysis confirms that input-level shift detection plays an important role in identifying distributional changes and providing warning signals, but also shows that input data shifts alone are not always predictive of downstream performance degradation.

Motivated by this observation, we further examine output-level signals and introduce a simple yet effective, \textbf{label-free} confidence-based performance degradation indicator. This indicator captures changes in the model's predictive confidence under shifting data conditions. The underlying intuition is that as a VLM operates farther from its reliable regime, its output confidence distribution becomes less separable and more uncertain which is a sign of performance degradation.

We evaluate the proposed framework using a large-scale pathology dataset spanning multiple medical sites with scanner-induced variability. Experimental results demonstrate that output-based confidence indicators closely track performance degradation, while input-level shift detection provides complementary diagnostic context. Together, these signals enable more reliable monitoring and interpretation of performance degradation in pathology VLMs under data shift.

The main contributions of this study are threefold:
\begin{itemize}
\item An in-depth investigation of data-shift-induced performance degradation in a state-of-the-art pathology VLM.

\item A label-free Confidence-based Degradation Indicator (CDI), which provides an effective signal for VLM degradation without requiring ground-truth labels.

\item DomainSAT, a lightweight GUI toolbox that integrates representative shift-detection methods and facilitates intuitive visualization of data shift patterns.
\end{itemize}

These components provide a practical framework for reliability monitoring of foundation models in digital pathology, with the DomainSAT toolbox publicly available at \url{https://github.com/guanharry/DomainSAT}.

\section{Related Work}

\subsection{Vision-Language Model}
Vision-Language Models (VLMs) learn joint representations of images and text, enabling zero-shot classification and retrieval. General-purpose VLMs such as CLIP~\cite{radford2021learning}, BLIP-2~\cite{li2023blip}, and LLaVA~\cite{liu2023visual} have demonstrated strong generalization across diverse visual domains.
Motivated by their success, several medical VLMs have been developed to improve clinical outcomes. These include MedCLIP~\cite{wang2022medclip}, BiomedCLIP~\cite{zhang2023biomedclip}, and LLaVA-Med~\cite{li2023llava}, which leverage paired medical image-text data to enhance medical understanding. 

In digital pathology, domain-specific VLMs have recently emerged to address the unique challenges of high-resolution histopathology images and fine-grained tissue semantics. Notably, PathGen-CLIP~\cite{PathGen}, the state-of-the-art model used in this study, is trained on 1.6 million pathology image-text pairs. PathGen-CLIP provides strong pathology-specific embeddings and has demonstrated very good performance in tumor classification tasks.
Despite rapid progress, the reliability and performance degradation of pathology VLMs under real-world data shift remain largely unexplored, motivating this study.

\subsection{Domain Shift Detection}
Domain (data) shift detection methods can generally be categorized into three groups: distance-based, statistic-based, and machine learning-based, as shown in Table~\ref{tab:algorithms}.

\subsubsection{Distance-Based Methods}
These methods quantify discrepancies between reference and target data using distance metrics, where larger values typically indicate stronger domain shift and increased risk of performance degradation. 
%
%
In~\cite{kore2024empirical}, a pre-trained autoencoder extracts latent features from chest X-rays, and MMD distance is computed to detect subtle distributional changes, including those associated with emerging conditions such as COVID-19. Stacke~\etal\cite{stacke2020measuring} use CNN-derived features and apply Wasserstein and KL divergences to assess latent-space shifts in medical image analysis. 

\begin{table}[t]
\caption{Representative Domain Shift Detection Algorithms Included in the DomainSAT Toolbox.}
\centering
\setlength{\tabcolsep}{3mm}{
\begin{tabular} {lllll} 
\toprule[1.2pt]

Category               &Algorithm          &Reference\\

\midrule

\multirow{5}*{Distance-Based}     
&MMD                          &~\cite{MMD} \\
&Wasserstein Distance         &~\cite{WD} \\
&Mahalanobis Distance         &~\cite{MD} \\
&JS Divergence                &~\cite{JSD} \\  
&KL Divergence                &~\cite{KLD} \\

\midrule

\multirow{5}*{Statistic-Based}    
&Kolmogorov-Smirnov test       &~\cite{KS} \\
&Wilcoxon rank-sum test        &~\cite{Wilcoxon} \\
&Cram\'{e}r-von Mises test    &~\cite{darling1957kolmogorov} \\
&$\chi^2$ test                 &~\cite{merkow2023chexstray} \\

\midrule

\multirow{4}*{Machine Learning-Based}    
&Domain Classifier             &~\cite{dreiseitl2022comparison} \\
&C2ST (Logistic Classifier)    &~\cite{lopez-paz2017revisiting} \\
&C2ST (Random Forest)          &~\cite{C2ST-RF} \\
&Autoencoder                   &~\cite{bobeda2023unsupervised} \\

\bottomrule[1.2pt]
\end{tabular}
}
\label{tab:algorithms}
\end{table}
\begin{figure*}[!tbp]
\setlength{\belowcaptionskip}{-2pt}
\setlength{\abovecaptionskip}{-2pt}
\setlength{\abovedisplayskip}{-2pt}
\setlength{\belowdisplayskip}{-2pt}
\center
 \includegraphics[width= 1.0\linewidth]{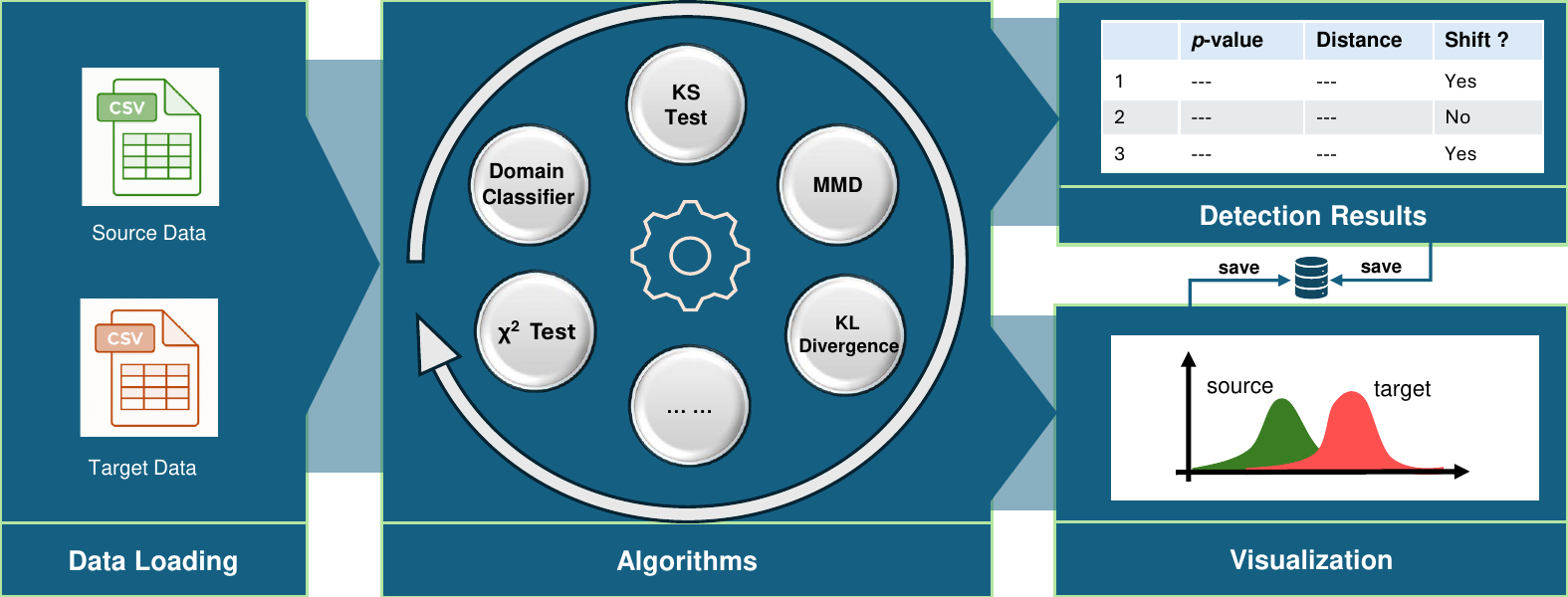}
 \caption{Overview of the DomainSAT workflow. The toolbox comprises three main components: (1) a data loading module for importing and preprocessing datasets; (2) an algorithm module that integrates multiple domain shift detection methods; and (3) an output module that generates detection results and visualizations.}
 \label{fig_workflow}
\end{figure*}

\subsubsection{Machine Learning-Based Methods}

A common strategy for shift detection is training a domain classifier (\eg, logistic regression) to distinguish source from target samples~\cite{dreiseitl2022comparison}. High classification accuracy indicates that the distributions differ significantly. This approach, known as the Classifier Two-Sample Test (C2ST)~\cite{lopez2017revisiting,kirchler2020two,pandeva2024evaluating}, is widely used for high-dimensional data. For example,~\cite{koch2024distribution} applies C2ST to monitor diabetic retinopathy screening, detecting shifts related to image quality, co-morbidities, and demographics.

\subsubsection{Statistic-Based Methods}
These methods apply hypothesis tests to compare source and target distributions, typically returning \emph{p}-values for interpretable decisions. They are often used on one-dimensional features to pinpoint sources of shift.
The Kolmogorov-Smirnov (KS) test~\cite{KS} is commonly used for data shift detection. CheXstray~\cite{merkow2023chexstray} employs KS test on DICOM metadata, image features, and model outputs for real-time monitoring. 

%

\section{Domain Shift Analysis Toolbox}
\subsection{Toolbox Design and Workflow}

\subsubsection{Overall}
As illustrated in Fig.~\ref{fig_workflow}, the DomainSAT (\textbf{Domain} \textbf{S}hift \textbf{A}nalysis \textbf{T}oolbox) consists of three primary modules: a) \emph{Data loading module}, which is responsible for importing both source and target datasets in a standardized format;
b) \emph{Algorithm module}, which integrates the classic domain shift detection methods, enabling users to perform in-depth analyses; 
c) \emph{Output module}, which produces interpretable shift detection results along with informative visualizations.

To maximize usability, the toolbox provides a \textbf{Graphical User Interface (GUI)}, enabling offline exploratory and retrospective analysis, visualization of data distributions, and domain shift assessment without requiring any coding.

\subsubsection{Workflow}
Using the GUI, users begin by uploading the source and target datasets (in \emph{.csv} format). 
They can then select one or more domain shift detection algorithms from the toolbox. After execution, DomainSAT generates detailed outputs including \emph{p}-values, distance metrics, or classifier scores, which can be saved as \emph{.csv} files for further analysis.
In addition, the toolbox provides visualization functions that display the distribution of each feature in both the source and target domains. These visual comparisons enable intuitive and in-depth feature-level analysis of domain shift.
\subsection{Algorithms}
The DomainSAT toolbox incorporates a bunch of popular domain shift detection algorithms, as detailed in Table~\ref{tab:algorithms}. These algorithms are grouped into three major categories according to their underlying mechanisms: distance-based methods, statistical testing methods, and machine learning-based methods.

Distance-based methods compute a numerical distance score that quantifies the discrepancy between source and target datasets. A domain shift is typically flagged when this score exceeds a predefined threshold.

Statistical testing methods evaluate whether the source and target data come from the same distribution. A small \emph{p}-value (typically below 0.05 or 0.02) indicates a statistically significant data shift.

Machine learning-based methods train a classifier to distinguish between source and target data. If the classifier achieves performance significantly above random chance (measured by metrics such as accuracy or Area Under the ROC Curve (AUC)), it indicates the presence of a detectable domain shift.

The toolbox and all included algorithms are implemented in Python. As a fully open-source project, users can freely extend it by adding and testing new algorithms following the existing pipeline. All implemented methods run cross-platform (\eg, Windows, MacOS) and do not require a GPU.

\begin{figure*}[ht]
\setlength{\belowcaptionskip}{-2pt}
\setlength{\abovecaptionskip}{-2pt}
\setlength{\abovedisplayskip}{-2pt}
\setlength{\belowdisplayskip}{-2pt}
\center
 \includegraphics[width= 1.0\linewidth]{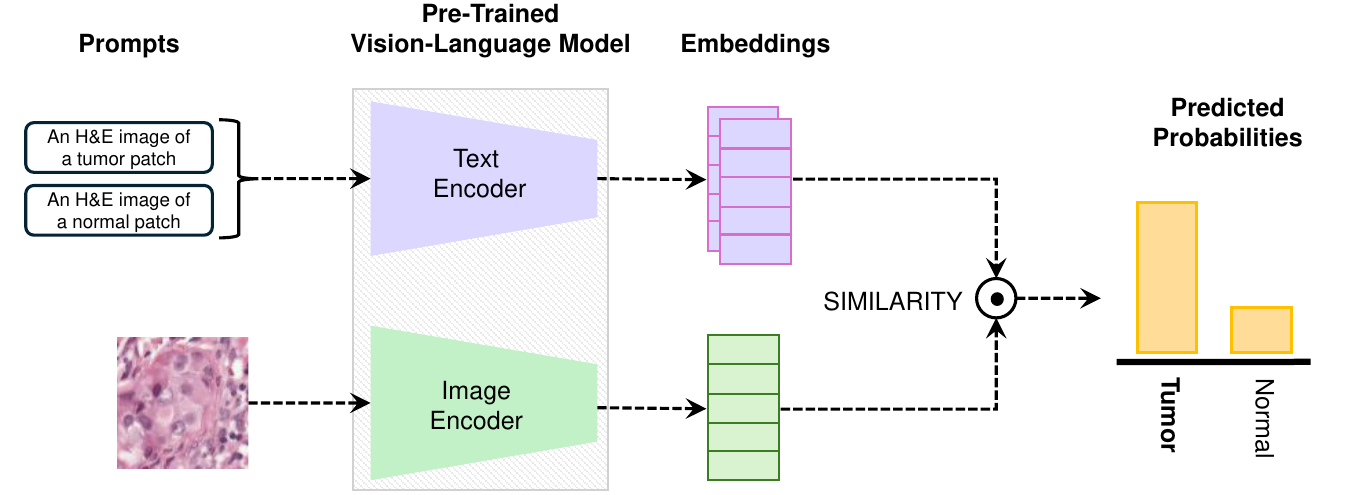}
 \caption{Illustration of the pretrained vision-language model (PathGen-CLIP) applied to histopathology image classification. The inputs are a histopathology image and two text prompts (``An H\&E image of a tumor patch" and ``An H\&E image of a normal patch"). The image and text are separately encoded and compared via similarity scoring. The resulting similarity scores are normalized into predicted probabilities for the two classes (tumor vs. normal). }
 \label{fig_CLIP}
\end{figure*}

\section{Explore Performance Degradation of Pathology VLM}
In this section, we study performance degradation of pathology VLMs under data shift by jointly analyzing input-level shift signals and output-level indicators, highlighting their complementary roles in reliable AI model monitoring.

\begin{table}[t]
\caption{Summary of the Camelyon17 sites used in this study. Sites 1–3 are combined as the in-distribution (ID) reference set, while Sites 4 and 5 serve as out-of-distribution (OOD) sets.}
\centering
\setlength{\tabcolsep}{2mm}{
\renewcommand{\arraystretch}{1.5}  
\begin{tabular} {lllll} 
\toprule[1.2pt]

Dataset       &Assignment         &Scanner    &Image Count   &Positives\\

\midrule

Site-1,2,3    &ID (Reference)    &Pannoramic       &33,560     &50\% \\

Site-4        &OOD-S1 (Unseen)        &Hamamatsu        &34,904     &50\% \\
Site-5        &OOD-S2 (Unseen)        &Philips          &85,054     &50\% \\

\bottomrule[1.2pt]
\end{tabular}
}
\label{tab:dataset}
\end{table}

\subsection{Dataset}
\subsubsection{Overview}
We use the WILDS Camelyon17 dataset, a publicly available histopathology benchmark designed to study domain shift~\cite{CAMELYON17,WILDS}. The dataset captures real-world data shift arising from differences in staining protocols and scanner types across five hospitals (Sites 1, 2, 3, 4, 5). Sites 1-3 share the same scanner for histopathology image acquisition, while Sites 4 and 5 each use a different scanner type. The dataset consists of image patches extracted from whole-slide pathological images, with patch-level binary labels indicating the presence or absence of tumor. A summary of the sites used in our experiments is provided in Table~\ref{tab:dataset}.

\subsubsection{Reference and OOD Sites}
For this study, we define:
\begin{itemize}
\item
In-distribution (ID) / reference site: Sites 1-3 are combined to simulate the deployment reference. This reflects the distribution available during model development and provides the baseline for monitoring.
\item
Out-of-Distribution (OOD) sites: Site 4 (OOD-S1) and Site 5 (OOD-S2) serve as unseen data, collected using different scanners and staining conditions. These mimic realistic domain shifts encountered after deployment.
\end{itemize}


\subsection{Model}
We evaluate PathGen-CLIP~\cite{PathGen}, a state-of-the-art pathology vision-language model for tumor classification that builds upon CLIP~\cite{CLIP} using a ViT-B/16 image encoder and a standard transformer-based text encoder. To simulate realistic post-deployment of AI products, the model is kept entirely frozen throughout all experiments, with no fine-tuning or parameter updates. The overall model architecture and inference pipeline are illustrated in Fig.~\ref{fig_CLIP}.

Importantly, PathGen-CLIP was \emph{not} pretrained or updated on Camelyon17, ensuring that all performance differences observed in our experiments stem solely from domain shift in the incoming data rather than from model adaptation or retraining. This avoids information leakage and more accurately reflects real-world monitoring scenarios, where the deployment data is entirely unseen to the AI model before.

In our setup, Sites 1, 2, and 3 of Camelyon17 are combined to serve as the simulated deployment reference dataset (Table~\ref{tab:dataset}), rather than using PathGen-CLIP's actual training data. This design mirrors practical conditions: the original training set of a commercial AI product is typically unavailable or inaccessible, so a portion of deployment data (on which the model exhibits expected performance) is used as the reference for ongoing monitoring.

\subsection{Inference}
For inference, we use the pathology VLM (PathGen-CLIP) to perform binary tumor classification with the label set $\mathcal{C}=\{\textbf{tumor},\textbf{normal}\}$.
We provide the model with two text prompts: ``An H\&E image of a tumor patch" and ``An H\&E image of a normal patch", which serve as class descriptions for the text branch of the VLM.

During inference, the model takes two inputs: 1) a pathology image patch, and 2) the text prompts.
The image encoder produces an image embedding $\mathbf{z}$, and the text encoder produces an embedding $\mathbf{t}_c$ for each class $c\in \mathcal{C}$.
The model then computes the similarity (correlation) between the image embedding and the embeddings of the two prompts. These similarity scores are converted into probabilities using a softmax function. The class with the higher probability is selected as the final prediction. For example, if the embedding of a pathology image patch is more similar to the embedding of ``An H\&E image of a tumor patch" than ``An H\&E image of a normal patch", then it is classified as tumor.

Mathmetically, we apply $\mathcal{L}_2$ normalization to both the image and text embeddings before computing similarity,
\[
\hat z=\frac{z}{\lVert z\rVert_2}, \qquad \hat t_c=\frac{t_c}{\lVert t_c\rVert_2},
\]
then the similarity scores are computed with the CLIP logit scale \(\alpha\),
\[
s_c=\alpha\,\hat z^{\top}\hat t_c,
\]
and class posteriors can be obtained as:
\[
p(y=c\mid x)=\frac{\exp(s_c)}{\sum_{c\in\mathcal{C}}\exp(s_{c})}, \quad c\in\mathcal{C}.
\]

For each pathology image, we record 1) the 512-dimensional normalized image embedding, and 2) the two-dimensional posterior probability vector over the two classes (tumor vs. normal). These stored posteriors are later used for output analysis.
\begin{figure*}[!tbp]
\setlength{\belowcaptionskip}{-2pt}
\setlength{\abovecaptionskip}{-2pt}
\setlength{\abovedisplayskip}{-2pt}
\setlength{\belowdisplayskip}{-2pt}
\center
 \includegraphics[width= 0.95\linewidth]{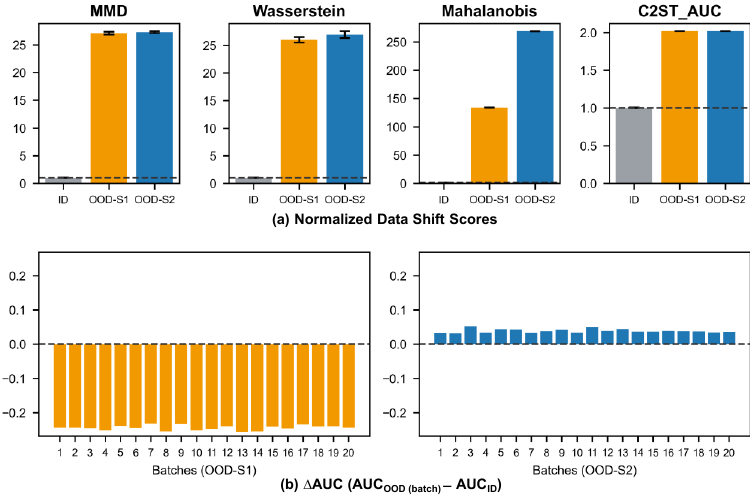}
 \caption{Input data shift and performance degradation of a state-of-the-art pathology VLM on out-of-distribution (OOD) sites.
(a) Data shift score (mean ± std) for 20 OOD subgroups (batches) from each OOD site, measured relative to the baseline of in-distribution (ID) dataset.
(b) Performance degradation (AUC drop) of the VLM across the 20 OOD batches; the baseline is the model's tumor-classification AUC on the full ID site.}
 \label{embedding_shift}
\end{figure*}
\subsection{Data Shift and Performance Degradation Analysis}
\subsubsection{Experimental Setting}
In this section, we analyze the data shift between the deployment site (ID) and the out-of-distribution sites (\ie, OOD-S1 and OOD-S2), and investigate how this shift relates to tumor classification performance of the pathology vision-language model.

\vspace{3pt}
\textbf{Shift Quantification Metric.}
To quantify data shift between the ID site and the OOD sites, we use four widely adopted metrics:
\emph{1) Maximum Mean Discrepancy (MMD)};
\emph{2) Wasserstein Distance};
\emph{3) Mahalanobis Distance}; and
\emph{4) the Classifier Two-Sample Test (C2ST)}, where the AUC of a logistic classifier is used as the shift score.


\vspace{3pt}
\textbf{Data Shift Baseline.}
Raw data shift scores are often difficult to interpret due to metric-dependent scales. To provide a meaningful reference, we construct an in-distribution (ID) data shift baseline. Specifically, we sample 20 ID batches (each with 5,000 samples and a balanced tumor/normal ratio) with replacement from the ID dataset and compute their shift scores relative to the full ID dataset using MMD, Wasserstein, Mahalanobis, and C2ST. The mean shift score across these ID batches is used as the baseline for each metric.

For each OOD batch, we compute its shift score relative to the full ID dataset and divide it by the corresponding baseline value. As a result, all reported shift scores indicate how many times larger the OOD shift is compared to ID baseline.
For example, if the baseline MMD is 5 and an OOD batch has an MMD of 100, the reported shift score is 20, meaning that the OOD shift is ten times larger than normal within-ID variation.

\vspace{3pt}
\textbf{Data Shift Computation of the OOD Sites.}
For OOD Site~1, we construct 20 out-of-distribution subgroups (batches), each with 5,000 samples (with a 50:50 tumor/normal ratio) by sampling from the full OOD site~1. 
For each batch \(i \in \{1, 2, \ldots, 20\}\) in OOD\text{-}S1, we compute its shift score (\ie, MMD, Wasserstein, Mahalanobis, C2ST) relative to the ID data shift score baseline.
The same procedure is applied to OOD-S2.


\vspace{3pt}
\textbf{Performance Degradation Quantification Metric.}
We first compute the performance of tumor classification (in terms of AUC) on the ID site, denoted as $\mathrm{AUC}_\mathrm{ID}$.
For Batch $i\in \{1, 2, ..., 20\}$ in OOD-S1, the AUC of tumor classification of the VLM is denoted as $\mathrm{AUC}_{i}$, then the performance degradation (on OOD Batch $i$) is defined as:
\begin{equation}
\mathrm{\Delta AUC}_i=\mathrm{AUC}_i-\mathrm{AUC}_\mathrm{ID}.
\end{equation}

Since OOD-S1 is divided into 20 subgroups (batches), the pathology VLM performs tumor classification on each batch. We compute the performance degradation for each subgroup (batch) and examine how it relates to the degree of data shift (\ie, its discrepancy from the ID dataset). The same procedure is applied to OOD-S2.


\subsubsection{Results}

Figure~\ref{embedding_shift}~(a) presents the data shift scores (mean ± std across 20 OOD batches) for OOD-S1 and OOD-S2. 
All OOD batches exhibit substantially higher shift scores compared to the baseline. 
MMD and Wasserstein show large fold increases (approximately $25\times$), while Mahalanobis distances are even larger (around $1.3\times10^{2}$ for OOD-S1 and $2.7\times10^{2}$ for OOD-S2). 
C2ST classification AUC also doubles relative to baseline, indicating that a simple classifier can readily distinguish OOD samples from ID samples.

Panel~(b) shows the performance degradation (AUC drop) of the pathology VLM relative to its in-distribution performance. 
For OOD-S1, the model's AUC consistently decreases across all 20 batches by approximately $0.25$. 
In contrast, for OOD-S2, the changes are small and slightly positive ($0.03\sim0.05$), indicating no meaningful degradation.


\subsubsection{Analysis}
All shift detection methods successfully distinguish the OOD datasets from the ID dataset, demonstrating their effectiveness in capturing appearance changes in pathology images. This confirms that input data shift detection provides valuable diagnostic information and can serve as an early warning signal of data distribution changes and potential performance variation.

Both OOD sites exhibit large data shifts; however, only OOD-S1 experiences substantial performance degradation. This observation highlights that input data shift is an \textbf{unsigned}, aggregate measure: it captures any change in data appearance (\eg, scanner or staining differences) without indicating whether such changes affect the model's performance.

As a result, a large input shift does not necessarily affect the VLM's classification boundary and may therefore represent a \emph{benign} shift. In contrast, some shifts can cause input samples to move closer to or overlap around the decision boundary, increasing uncertainty and misclassification and leading to significant performance degradation, representing a \emph{harmful} shift.
Therefore, input data shift detection serves as a useful \textbf{\emph{diagnostic}} signal, indicating that ``something has changed", but it is not, by itself, a reliable \textbf{\emph{prognostic}} indicator of whether harmful degradation in AI performance will occur.

\subsection{Output-based Indicator for Performance Degradation}

\begin{figure}[t]
\setlength{\belowcaptionskip}{-2pt}
\setlength{\abovecaptionskip}{-2pt}
\setlength{\abovedisplayskip}{-2pt}
\setlength{\belowdisplayskip}{-2pt}
\center
 \includegraphics[width= 1.0\linewidth]{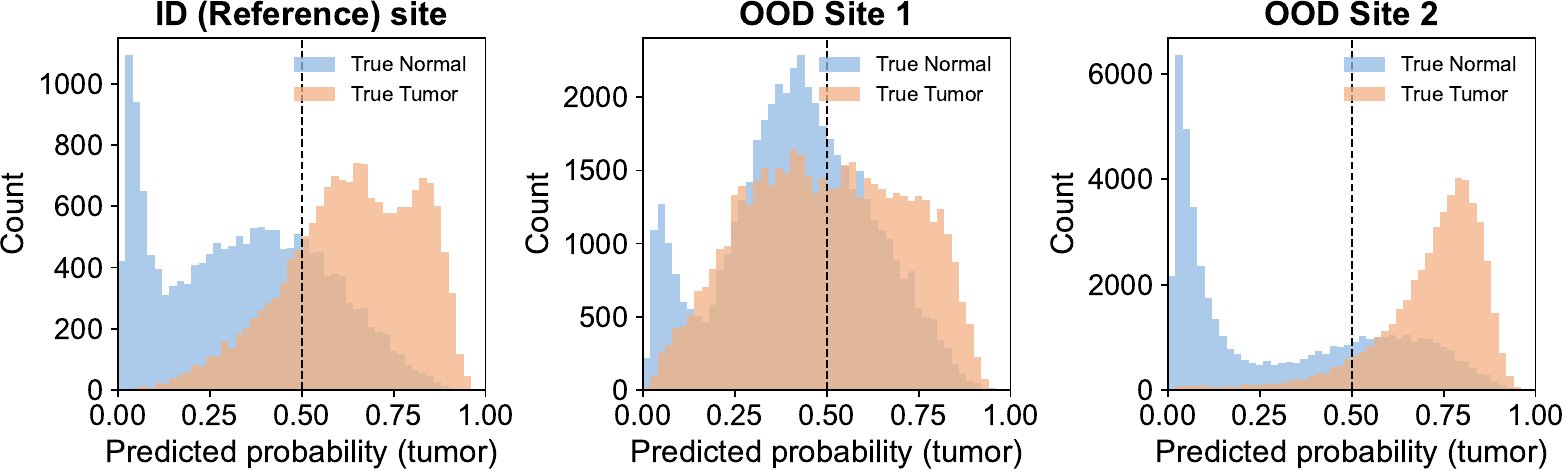}
 \caption{Predicted tumor probability distributions on the in-distribution (ID) site and two out-of-distribution (OOD) sites. Each panel shows histograms of predicted tumor probabilities for true normal (blue) and true tumor (orange) samples. On OOD Site 1, predictions collapse toward the decision boundary ($0.5$). On OOD Site 2, predictions are more separated.}
 \label{p_distributions}
\end{figure}

\subsubsection{Motivation}

From the above experimental study, we observe that input data shifts do not always lead to performance degradation. 
Although detecting a large data shift can be useful for triggering an alarm for potential AI degradation, a more intuitive strategy is to directly monitor and detect degradation itself.
Our experiments suggest that output-based variations are more closely and directly related to performance changes, motivating us to design output-based degradation indicators.
To illustrate this motivation, we first visualize the outputs of the pathology VLM.

For each input pathology image (patch), the pathology VLM produces two probabilities:  
1) the probability that the image is a tumor image, denoted as $p(\text{tumor})$;  
2) the probability that the image is a normal (healthy) image, denoted as $p(\text{normal})$.  
For an input \textbf{tumor} image, if the VLM performs well, $p(\text{tumor})$ should be high (\eg, $0.8$), while $p(\text{normal})$ should be low (\eg, $0.2$).  
Following this intuition, we visualize $p(\text{tumor})$ for all images in the ID site and both OOD sites.

Figure~\ref{p_distributions} shows the distribution of $p(\text{tumor})$ on the ID site and on the two OOD sites, \ie, OOD-S1 and OOD-S2.  
The blue histogram represents the distribution of $p(\text{tumor})$ for normal (healthy) pathology images, whereas the orange histogram represents the distribution of $p(\text{tumor})$ for tumor images.
From Figure~\ref{p_distributions}, we can see that on the ID site the two distributions are well separated (normals concentrated near $0$, tumors near $1$). 
This indicates that for true normal images, the predicted tumor probability is low, while for true tumor images, the predicted tumor probability is high. This is precisely the behavior expected from a well-performing pathology AI model.

On OOD-S1, however, the two distributions exhibit substantial overlap. 
In this setting, instead of assigning a confident and polarized probability (\eg, $80\%$ tumor vs.\ $20\%$ normal), the VLM often produces more ambiguous estimates (\eg, $47\%$ tumor vs.\ $53\%$ normal). 
As a result, more true tumor images are misclassified as normal and vice versa, leading to a significant deterioration in performance.
On OOD-S2, the situation is quite different. 
Although the input data shift is very large (as quantified in the previous section and shown in Figure~\ref{embedding_shift}), the output distributions (of $p(\text{tumor})$) for true normals and tumors remain well separated. 
Consequently, the model's performance (as shown in Figure~\ref{embedding_shift}) does not exhibit meaningful degradation.

Taken together, these patterns suggest that \emph{output-based} indicators are more sensitive and directly informative for detecting performance degradation in medical AI systems.


\subsubsection{Output-Based Degradation Indicator}

When a pathology VLM is applied to an out-of-distribution dataset $\mathcal{S}$, let $p_i \in [0,1]$ denote the predicted probability of the positive class (tumor) for sample $i$. 
Thus, the model outputs a set of predicted probabilities $\{p_i\}_{i=1}^n$ for the $n$ samples in $\mathcal{S}$. 
To detect potential performance degradation on $\mathcal{S}$, we introduce an output-based indicator called the \textbf{Confidence-based Degradation Index (CDI)}. 
We instantiate CDI in two simple and label-free forms: a \emph{Margin-based CDI (CDI\_M)} and an \emph{Entropy-based CDI (CDI\_H)}.

\vspace{3pt}

\textbf{Margin-based CDI (CDI\_M)}.
This indicator measures how close the model’s predictions are, on average, to the decision boundary $\mathcal{P}$ (\ie, $p_i > \mathcal{P}$ indicates tumor; otherwise normal).  
For a dataset $\mathcal{S}$, we define

\begin{equation}
\mathrm{CDI\_M}(\mathcal{S}) = 2\,\frac{1}{n}\sum_{i=1}^n \bigl|p_i - \mathcal{P}\bigr| \\
\end{equation}
In our setting, we set the decision boundary $\mathcal{P}$ to $0.5$ because our task is a binary classification problem with approximately balanced positive and negative classes.
A value of $\mathrm{CDI\_M}=0$ corresponds to very low confident predictions (\ie, $p_i \approx 0.5$), whereas values approaching $1$ indicate that predictions cluster near $0$ and $1$.
Thus, as $\mathrm{CDI\_M}$ decreases ($\mathrm{CDI\_M}\!\downarrow$), confidence decreases, which can signal potential performance degradation ($\mathrm{AUC}\!\downarrow$).

To detect performance degradation of a medical AI model on a dataset $\mathcal{S}_{i}$ relative to a reference dataset $\mathcal{S}_{\mathrm{ref}}$, we compute
\begin{equation}
\Delta \mathrm{CDI\_M} 
= \mathrm{CDI\_M}(\mathcal{S}_{\mathrm{i}}) - \mathrm{CDI\_M}(\mathcal{S}_{\mathrm{ref}}).
\end{equation}
A \textbf{negative} $\Delta \mathrm{CDI\_M}$ indicates a confidence collapse, often associated with performance degradation, whereas a \textbf{positive} value suggests more confident predictions and no degradation.

\vspace{3pt}
\textbf{Entropy-based CDI (CDI\_H).} 
This indicator captures output ambiguity via the mean binary Shannon entropy~\cite{lin2002divergence}.  
Let the Shannon entropy (with the logarithm base set to $2$) for a predicted probability $p$ be
\[
H(p) = -\,p\log p \;-\; (1-p)\log(1-p).
\]
For a dataset $\mathcal{S}$ with $n$ samples, we define
\begin{equation}
\mathrm{CDI\_H}(\mathcal{S}) = 1 - \frac{1}{n}\sum_{i=1}^{n} H(p_i).
\end{equation}
The value $\mathrm{CDI\_H}$ reflects the average prediction confidence in the dataset $\mathcal{S}$. 
It equals 1 (its maximum) when predictions are fully confident (\ie, $p=0$ or $p=1$), and decreases toward 0 as predictions move closer to the decision boundary (\ie, $p=0.5$). 
Thus, as the $\mathrm{CDI\_H}$ decreases ($\mathrm{CDI\_H}\!\downarrow$), confidence declines, which may indicate potential performance degradation ($\mathrm{AUC}\!\downarrow$).

To detect performance degradation of a medical AI model on a dataset $\mathcal{S}_{i}$ relative to a reference dataset $\mathcal{S}_{\mathrm{ref}}$, we compute
\begin{equation}
\Delta \mathrm{CDI\_H} 
= \mathrm{CDI\_H}(\mathcal{S}_{\mathrm{i}}) - \mathrm{CDI\_H}(\mathcal{S}_{\mathrm{ref}}).
\end{equation}
A \textbf{negative} $\Delta \mathrm{CDI\_H}$ reflects a collapse in confidence, which is often linked to performance degradation, whereas a \textbf{positive} value indicates stronger confidence without degradation.

\begin{figure}[t]
\setlength{\belowcaptionskip}{-2pt}
\setlength{\abovecaptionskip}{-2pt}
\setlength{\abovedisplayskip}{-2pt}
\setlength{\belowdisplayskip}{-2pt}
\center
 \includegraphics[width= 1.0\linewidth]{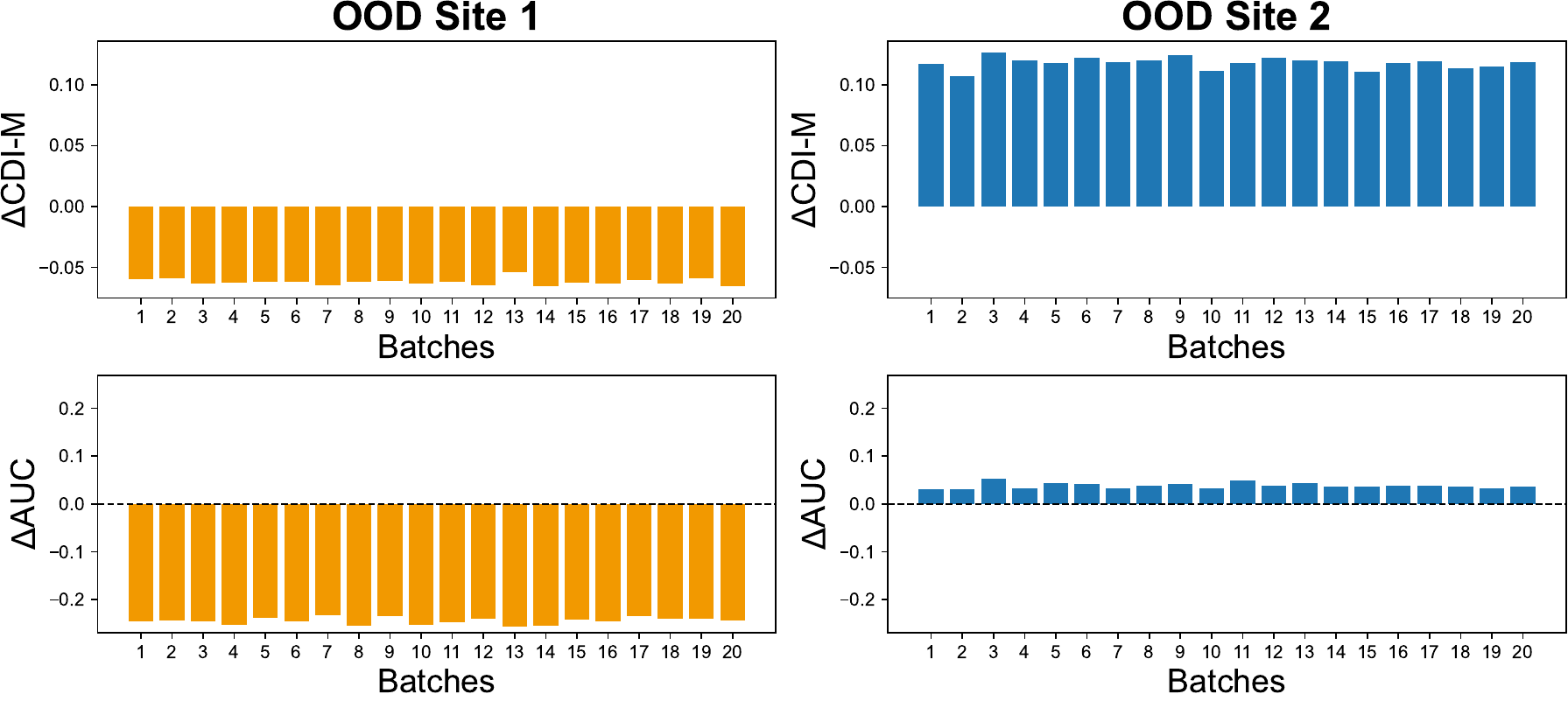}
 \caption{Subgroup (Batch)-wise changes in confidence ($\Delta\mathrm{CDI\_M}$, top) and performance ($\Delta\mathrm{AUC}$, bottom) for OOD-S1 and OOD-S2. OOD-S1 shows confidence collapse with large AUC drops, while OOD-S2 shows increased confidence with stable or slightly improved AUC.}
 \label{CCI-M}
\end{figure}
\begin{figure}[!tbp]
\setlength{\belowcaptionskip}{-2pt}
\setlength{\abovecaptionskip}{-2pt}
\setlength{\abovedisplayskip}{-2pt}
\setlength{\belowdisplayskip}{-2pt}
\center
 \includegraphics[width= 1.0\linewidth]{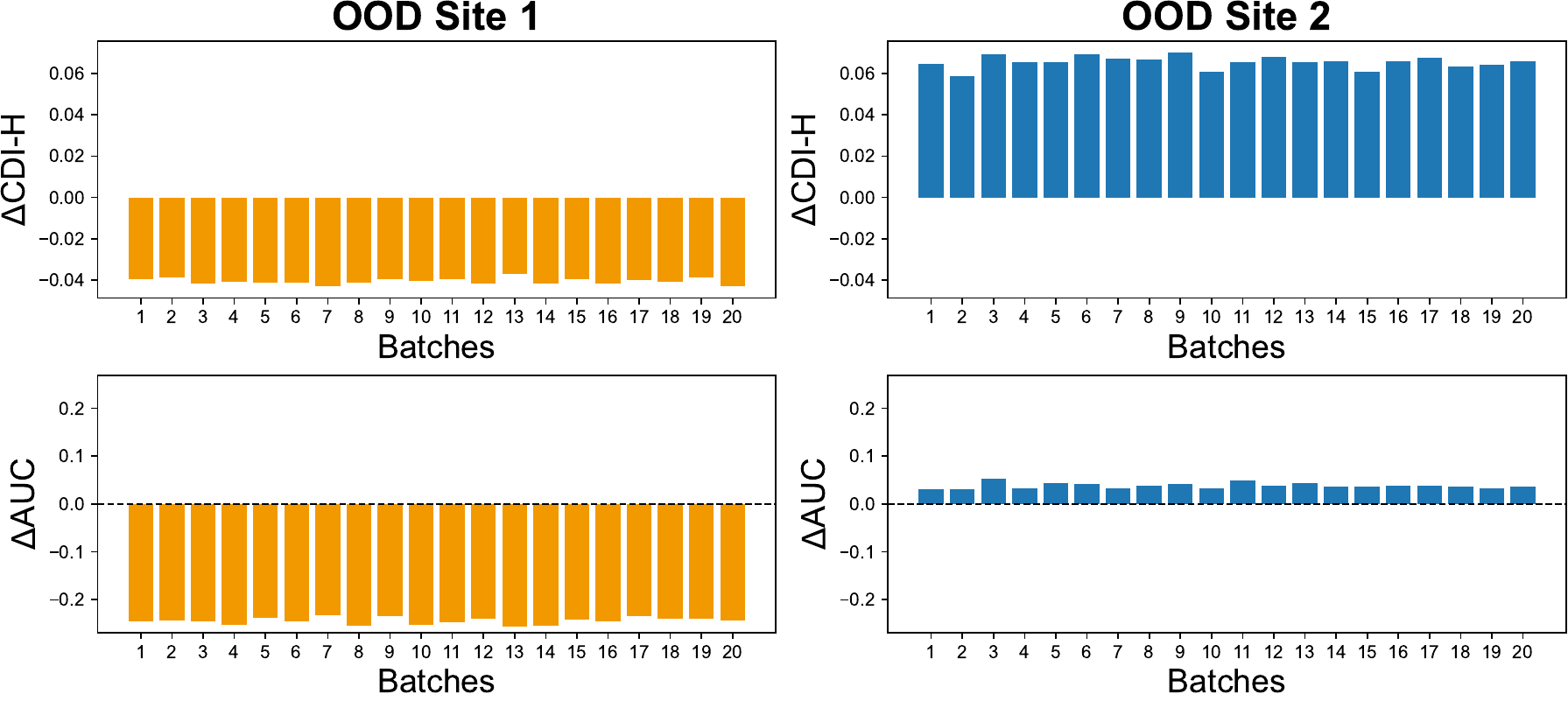}
 \caption{Subgroup (batch)-wise changes in confidence ($\Delta\mathrm{CDI\_H}$, top) and performance ($\Delta\mathrm{AUC}$, bottom) for OOD-S1 and OOD-S2. OOD-S1 exhibits confidence collapse with significant AUC drops, whereas OOD-S2 shows increased confidence accompanied by minor improved AUC.}
 \label{CCI-H}
\end{figure}
\subsubsection{Results}
For OOD-S1, we build 20 out-of-distribution subgroups (batches), each with 5,000 samples (with a 50:50 tumor/normal ratio) by sampling from the full OOD-S1. 
For each batch \(i \in \{1, 2, \ldots, 20\}\) in OOD-S1, we compute its $\Delta\mathrm{CDI\_M}_i$ and $\Delta\mathrm{CDI\_H}_i$, and the variation of AUC (\ie, $\Delta\mathrm{AUC}_i$).
The same procedure is applied to OOD-S2.

Figure~\ref{CCI-M} shows the $\Delta\mathrm{CDI\_M}$ (top row) and the corresponding $\Delta\mathrm{AUC}$ (bottom row) in all the 20 subgroups (batches) of each OOD sites.
The two sites share identical y-axis limits so that magnitudes are directly comparable.

On \textbf{OOD-S1}, every batch shows a negative $\Delta \mathrm{CDI\_M}$ ($\approx -0.05$ to $-0.06$), indicating that predictions shift toward the decision boundary with increased uncertainty and loss of confidence. At the same time, $\Delta \mathrm{AUC}$ is consistently negative and large ($-0.23$ to $-0.25$), demonstrating substantial performance degradation.

In contrast, on \textbf{OOD-S2}, all batches have $\Delta \mathrm{CDI\_M} > 0$ ($\approx +0.09$ to $+0.12$), indicating that predictions move slightly \emph{farther} from decision boundary relative to ID, reflecting increased confidence. The corresponding $\Delta \mathrm{AUC}$ values are small and slightly positive (about $+0.03$ to $+0.06$), suggesting a minor performance gain.

Overall, changes in confidence show close relationship with the performance degradation: confidence collapses ($\Delta\mathrm{CDI\_M}<0$) then performance decreases ($\mathrm{AUC}\!\downarrow$), whereas confidence does not collapse ($\Delta\mathrm{CDI\_M}>0$), performance keeps stable or slightly improved.

Figure~\ref{CCI-H} shows the $\Delta \mathrm{{CDI}\_H}$ alongside $\Delta \mathrm{AUC}$ across OOD sites. On \textbf{OOD-S1}, $\Delta \mathrm{CDI\_H}$ is consistently negative while AUC drops sharply, whereas on \textbf{OOD-S2}, $\Delta \mathrm{CDI\_H}$ is positive while performance (AUC) remains stable or slightly improved. These patterns mirror those of $\mathrm{CDI\_M}$, confirming that this indicator aligns closely with performance changes.
\subsubsection{Discussion}

Across the two OOD sites, the output-space indicators and performance change show close relationship, consistent with the probability histograms. 
When \textbf{Confidence-based Degradation Index} decreases, performance declines; when \textbf{Confidence-based Degradation Index} remains intact, performance stays stable or even shows slight improvement. 
This demonstrates that output-space indicators provide meaningful \emph{prognostic} information: when predictions drift toward indecision, degradation tends to follow, whereas lower ambiguity corresponds to stable or improved performance.

These findings suggest a clear division of roles between input-shift detection and output-based indicators. 
Input data shift detection tells us that “something has changed”, but does not reveal whether the change will harm the pathology VLM’s performance. 
Output-based indicators, in contrast, help determine whether the model is experiencing \emph{actual} degradation.

In practice, these two approaches are complementary.
Input shift detection is well suited for raising early warning signals, while output-based indicators, though involving slightly delayed processing (than the input data), can provide a more direct alert of performance degradation.
Together, these signals can be combined for more reliable medical AI monitoring.



\section{Conclusion}
This study investigates performance degradation of a state-of-the-art pathology vision-language model under data shift and examines both input-level data shift detection and output-based indicators for reliability monitoring. We show that input data shift detection provides valuable diagnostic information by identifying distributional changes and serving as an early warning signal, but that such shifts do not always translate into performance degradation. To complement input-level monitoring, we introduce a lightweight, label-free output-based confidence indicator that more directly reflects degradation in model performance under data shift. We also develope DomainSAT, an open-source GUI toolbox that enables systematic shift analysis and intuitive visualization of feature-space changes. Together, these complementary input- and output-level approaches offer practical tools and insights for reliable monitoring of pathology vision-language models in real-world deployment settings.



\section*{Acknowledgment}
This work was partially supported by the National Library of Medicine under Grant No. 1R01LM014239.


\bibliographystyle{IEEEtran}
\bibliography{mybib}
\end{document}